\documentclass[runningheads]{llncs}
\usepackage[T1]{fontenc}
\usepackage{multirow} 
\usepackage{graphicx}
\usepackage{xcolor}
\usepackage{soul}
\usepackage{caption}
\usepackage{overpic}
\usepackage{adjustbox}
\usepackage{subcaption}
\usepackage{float}
\usepackage{booktabs}

\usepackage[utf8]{inputenc} 

\usepackage{cite}

\begin{document}
\titlerunning{Toward Better Optimization of Low-Dose CT Enhancement} 
\title{Toward Better Optimization of Low-Dose CT Enhancement: A Critical Analysis of Loss Functions and Image Quality Assessment Metrics}



%
%
\author{Taifour Yousra \and
Beghdadi Azeddine \and
Marie Luong\and
Zuheng Ming}
\authorrunning{Y. Taifour et al.}

%
\institute{L2TI, University Sorbonne Paris Nord, UR 3043, Villetaneuse, F-93430, France }
\maketitle              
\begin{abstract}
Low-dose CT (LDCT) imaging is widely used to reduce radiation exposure to mitigate high exposure side effects, but often suffers from noise and artifacts that affect diagnostic accuracy. To tackle this issue, deep learning
models have been developed to enhance LDCT images. Various loss
functions have been employed, including classical approaches such as Mean Square Error and adversarial losses, as well as customized loss functions(LFs) designed for specific architectures. Although these models achieve remarkable performance in terms of PSNR and SSIM, these metrics are limited in their ability to reflect perceptual quality, especially for medical images. In this paper,
we focus on one of the most critical elements of DL-based architectures, namely the loss function. We conduct an objective analysis of the relevance of different loss-functions for LDCT image quality enhancement and their consistency with image quality metrics. Our findings reveal inconsistencies between LFs and quality metrics, and highlight the need of consideration of image quality metrics when developing a new loss function for image quality enhancement.
\end{abstract}

\section{Introduction}
Medical imaging-guided diagnosis has opened broad avenues for disease detection, patient treatment, and improving survival outcomes. Among various medical imaging modalities, Computed Tomography (CT) stands out for its ability to provide highly detailed information about internal anatomical structures, surpassing in-plane imaging techniques (e.g., X-ray). In addition, it plays a crucial role in optimizing treatment doses during image-guided cancer treatment, particularly in the case of image-guided radiotherapy. However, acquiring high-quality diagnostic CT images often requires exposing patients to significant doses of ionizing radiation, which increases the risk of the development of secondary cancer or DNA damage. Low-Dose CT (LDCT) imaging has been introduced to reduce radiation exposure while still maintaining diagnostic capacity. Nevertheless, radiation dose levels directly affect image quality. 
As the radiation dose decreases, the image quality also decreases with higher noise visibility and low-contrast structures, making the perception of fine anatomical details and lesions more difficult. Therefore, enhancing the quality of LDCT images is essential to ensure the reliability and accuracy of radiologists' diagnoses.

In this context, deep learning solutions have shown remarkable performance in reducing noise and recovering structural details from LDCT images \cite{kulathilake2023review}. State-of-the-art approaches include convolutional neural networks \cite{chen2017low}, UNet-based models \cite{zhang2023novel} and some variants \cite{xiong2024re,mazandarani2023unext}. Generative Adversarial Network  \cite{yang2018low} have been used to produce more realistic CT images, enhancing the similarity between low-dose and normal dose CT scans. LitFormer\cite{chen2024lit} and CT-Former\cite{wang2023ctformer}, transformer based models implement attention mechanisms to better capture and enhance relevant regions quality. Furthermore, diffusion models such as CocoDiff\cite{gao2022cocodiff} and DDPM\cite{xia2022low} iteratively denoise the image to obtain an acceptable image quality. Recently, authors in \cite{ozturk2024denomamba} and \cite{li2024ct}, showed that incorporating Mamba into the LDCT enhancement model significantly improves LDCT image quality and excels in all the previously discussed approaches. Although these architectures achieved high PSNR and SSIM scores in distortion removal, they do not necessarily guarantee improved visual quality of the relevant structures. 

Given the pivotal role of the loss function (LF) in guiding the model's learning process, we investigate various LFs dedicated to reconstruction tasks and analyze how they influence the relationship between model learning and image quality. To assess this relationship, we analyzed the performance improvement throughout the training process to identify monotonic behavior. The evaluation was conducted using a combination of full reference (FR) and no reference(NR) quality metrics, carefully selected to estimate both the level of distortion and perceptual aspects of image quality, such as noise, sharpness and contrast.
The main contributions of this paper are summarized below:
\begin{itemize}
\item provide a comprehensive evaluation framework for LFs used in image quality enhancement training models
\item highlight the interest of integrating quality measurement into the development of LFs to avoid any inconsistencies between them.
\item open new perspectives for the development of more consistent and quality-guided LFs in order to optimize deep learning-based models for image quality enhancement.
\end{itemize}

\section{Related works}
In deep learning-based solutions for inverse problems, such as LDCT recontruction, given a set of \( \mathcal{N} \) training pairs \( \{ (x_i, y_i) \}_{i=1}^N \), the goal is to learn optimal network weights \( \theta^* \) of a deep learning model represented as a function \( \phi_\theta : \mathcal{X} \rightarrow \mathcal{Y} \) that minimizes the discrepancy between the model-enhanced  CT image \( \phi_\theta(x_i) \) and the normal-dose CT  image \( y_i \), based on a predefined loss function \( \mathcal{L} \). Formally, the optimal parameters \( \theta^* \) are obtained by solving the following optimazation problem:
\[
\theta^* = \arg\min_{\theta} \frac{1}{N} \sum_{i=1}^{N} \mathcal{L} \big( \phi_\theta(x_i), y_i \big).
\]
This process allows the model to infer an approximate inverse operator that recovers normal dose CT images \( y_i \) from low dose CT samples \( x_i \).

Various loss functions (LFs) have been proposed in the context of LDCT image quality enhancement to reduce noise while preserving fine structural details crucial for accurate diagnosis. Among them, Mean Squared Error (MSE) remains a baseline objective, broadly used across diverse architectures, including convolutional networks \cite{chen2017low}, transformers \cite{wang2023ctformer}, and GAN-based models\cite{huang2021gan}. However, as a pixel-wise metric, MSE tends to introduce over-smoothing effects, often suppressing important diagnostic details along with noise. To mitigate this issue, complementary losses are often combined with MSE to better balance noise reduction and structure preservation.

For instance, Liang et al.~\cite{liang2020edcnn} introduced an edge-preserving block with a compound loss combining MSE LF and multiscale perceptual LF, based on features extracted from a pre-trained ResNet to address issues such as over-smoothing and detail preservation in the denoised images. Similarly, \cite{gholizadeh2020deep} and \cite{ge2020adaptive} incorporated VGG-based perceptual LFs to enhance perceptual quality.

In parallel, adversarial learning approaches \cite{liu2023solving,huang2021gan} have shown promising results in generating visually realistic outputs, though often at the cost of underestimating residual noise. To address this, WGAN-VGG \cite{li2021low} incorporates a VGG-based perceptual loss into the adversarial framework, improving structure preservation. DUGAN \cite{huang2021gan} further enhances structural fidelity by combining two losses in both image and gradient domains. Additionally, Wang et al. \cite{wang2023self} proposed a hybrid LF, introducing a weighted patch loss (WPLoss) that adapts to spatially varying noise levels, alongside a high-frequency loss designed to restore textures and details typically lost with MSE alone.

More recent transformer-based models also address both denoising and structure enhancement. LIT-Former \cite{chen2024lit}, for instance, leverages self-attention mechanisms to simultaneously denoise and deblur, trained with a dual-purpose LF tailored to optimize both tasks. Likewise, ASCON \cite{chen2023ascon} integrates anatomical semantics into the denoising process, combining a global MSE for pixel fidelity with a contrastive loss that enforces anatomical consistency by learning relationships between positive and negative pairs though this comes at the cost of increased model complexity and longer training.

Emerging diffusion-based approaches \cite{gao2022cocodiff,xia2022low} offer a promising alternative, refining noise through iterative processes and relying on $\ell_2$ loss function. However, these methods raise computational concerns due to their slow convergence and limited adaptability to complex medical images. Recently, DenoMamba \cite{ozturk2024denomamba} introduced a fused state-space model to capture long-range dependencies in CT images, trained with an $\ell_1$ loss, achieving state-of-the-art PSNR and SSIM scores. In parallel, Chen et al. \cite{chen2024low} explored the use of large language models (LLMs) to align low dose and normal dose CT images in continuous perceptual and discrete semantic spaces. Yet, despite their remarkable performance, SOTA methods fail to balance between denoising and enhancing the visibility of clinically significant structures, such as tumors critical for image guided diagnosis.

Unlike image quality assessment (IQA) in the common context, where we seek to measure a distance between the observed image and the associated reference or a priori information about this reference, the assessment of the level of quality enhancement is different \cite{le1992image}. In fact, enhancement can produce unpredictable and uncontrollable side effects, which makes the assessment complex \cite{beghdadi2020critical}. A few metrics have been proposed for measuring the level of image quality enhancement. Nevertheless, in this study we use the quality metrics defined in the general context of IQA.
Two approaches are considered: FR quality assessment, when a reference image is available, and NR quality assessment, in the absence of reference. FR-IQAs evaluate the quality of the enhanced image with respect to a target reference image. PSNR and SSIM are the most widely adopted metrics to assess LDCT enhancement due to their simplicity. However, due to the complexity of medical images and the necessity of preserving fine anatomical structures for accurate diagnosis, conventional metrics are often inadequate for fully capturing perceptual quality.

To address these limitations, perceptual FR metrics such as LPIPS
\cite{zhang2018unreasonable},
VIF and DISTS\cite{9298952} evaluate more meaningful structural and perceptual similarities, going beyond simple pixel-wise dissimilarities. In contrast, in real clinical-world scenarios, high-quality reference images may not be available, making NR IQA metrics more suitable. BRISQUE\cite{moorthy2011blind} and NIQE\cite{mittal2012making}, which estimate quality based on statistical models of natural images are the most used. More recent approaches, such as CLIP-IQA\cite{wang2023exploring} uses large language models (CLIP) to assess quality based on learned perceptual and semantic relationships, enabling better alignment with human visual preferences.

Despite their widespread use, a significant gap exists between the LFs employed to train LDCT enhancement models and the IQA metrics used to assess the models. LFs such as MSE, adversarial, and perceptual losses guide model optimization during training, leading to strong performance on commonly used metrics like PSNR and SSIM. However, these models often fail to preserve critical structures and fine details necessary for accurate image-guided diagnosis. This highlights the need to explore the consistency between conventional LFs used for enhancement tasks and IQA metrics, especially NR metrics, which fits clinical requirements.

\section{Materials and Methods }
In this study, we evaluate the performance of commonly adopted LFs in image reconstruction and enhancement models by analyzing their coherence with IQA measures in the context of LDCT image enhancement. We investigate how optimizing a LF within a given architecture relates to IQA outcomes. To this end, we select a representative LDCT enhancement model \cite{ozturk2024denomamba}, implement and analyze it under various LFs, then we evaluate the enhanced images using a set of FR and NR IQAs. 
All experiments were conducted using a single NVIDIA A100 GPU with 40~GiB of memory to ensure efficient processing and rapid computation of the enhancement model~\cite{ozturk2024denomamba}. The model was trained using the Adam optimizer with an initial learning rate of \(1 \times 10^{-4}\), which was adjusted during training via a scheduler based on validation loss. A batch size of 1 was used in all experiments, and validation was performed every 5 epochs while handling overfitting based on training and validation losses. To assess the effectiveness of different loss functions, the model was trained for 25, 60, 50, and 40 epochs using Charbonnier, L1, MSE, and VGG loss functions, respectively.

\subsection{Loss Functions}
We considered LFs commonly used in image quality enhancement tasks. Table~\ref{summary_loss} groups these functions into pixel-based and feature-based categories. Their effectiveness is assessed within the context of a deep learning-based LDCT image enhancement pipeline~\cite{ozturk2024denomamba}. To better analyze the consistency and impact of each loss function, we evaluate model performance at five training stages, with interval gaps selected to reflect meaningful learning progress and training efficiency.
\begin{table}[ht]
\centering
\begin{tabular}{|c|c|l|}
\hline
\textbf{Loss Category} & \textbf{Loss Function} & \textbf{Equation} \\
\hline
\multirow{3}{*}{Pixel-based Loss} 
  & L1 Loss & \( \mathcal{L}_1 = \frac{1}{N} \sum_{i=1}^N |x_i - \hat{x}_i| \) \\
\cline{2-3}
  & Mean Squared Error (MSE) & \( \mathcal{L}_{MSE} = \frac{1}{N} \sum_{i=1}^N (x_i - \hat{x}_i)^2 \) \\
\cline{2-3}
  & Charbonnier Loss & \( \mathcal{L}_{Charb} = \sum_{i=1}^N \sqrt{(x_i - \hat{x}_i)^2 + \epsilon^2} \) \\
\hline
Feature-based Loss & VGG Perceptual Loss & \( \mathcal{L}_{VGG} = \sum_{k} \left\| \phi_k(x) - \phi_k(\hat{x}) \right\|_2^2 \) \\
\hline
\end{tabular}
\caption{Summary of Loss Functions considered in this study}
\label{summary_loss}
\end{table}
\subsection{Image Quality assessment metrics}
It should be noted that in this experiment we are interested in measuring the level of improvement in image quality at different stages, i.e. epochs, of the learning process.  To do this, we consider two categories of image quality metrics, namely Full-reference (FR) and No-reference (NR).
FR metrics are computed between normal-dose CT (NDCT), considered as the reference, and enhanced LDCT images. While in the case of NR metrics, which reflect real clinical scenarios where reference images are unavailable, only the image resulting from the enhancement process is assessed. The FR metrics used in this experiment are PSNR, SSIM\cite{wang2004image}, VIF  \cite{sheikh2005visual},  LPIPS\cite{zhang2018unreasonable} and its variant ST-LPIPS\cite{ghildyal2022stlpips}, designed to capture perceptual differences while being robust to small spatial shifts and imperceptible changes, and DISTS\cite{9298952}, sensitive to texture and structure preservation. For blind assessment, 7 representative NR metrics covering different notions and concepts related to perceptual aspects and statistical characterization of scenes are used, namely BRISQUE\cite{moorthy2011blind}, NIQE\cite{mittal2012making}, AHIQ \cite{9857006} DBcnn\cite{zhang2020blind}, ARNIQA\cite{agnolucci2024arniqa}, MANIQA\cite{yang2022maniqa}, and finally, CLIP-IQA\cite{wang2022exploring} based on large language models (LLM) to assess both perceptual and semantic distortions.
\subsection{Dataset}
Since our objective is to analyze the consistency between LF and the considered image quality metrics in the context of LDCT image enhancement, we use the publicly available Mayo2016 dataset, provided by NIH Mayo Clinic AAPM  \cite{mccollough2016tu}. The dataset consists of normal dose CT images from 10 patients, acquired at a tube voltage of 120 kV and a tube current of 200 mAs. To simulate the effect of dose reduction, quarter dose CT pairs were generated by introducing Poisson noise in the projection domain. For training, we selected normal and low-dose CT pairs from 8 patients of 512 × 512 resolution and employed a 5-fold leave-patients-out cross-validation strategy, ensuring that each patient contributed to both training and testing.
\section{Results and discussion}
\begin{figure}[h] 
    \centering
    \begin{subfigure}{0.4\textwidth}
        \centering
        \includegraphics[width=\linewidth]{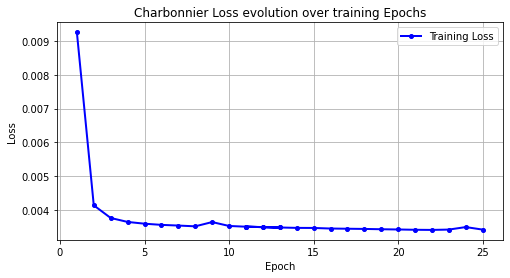}
        \caption{Charbonnier Loss}
    \end{subfigure}
    \hspace{0.03\textwidth}
    \begin{subfigure}{0.4\textwidth}
        \centering
        \includegraphics[width=\linewidth]{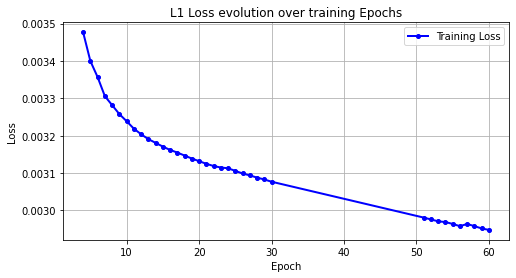}
        \caption{L1 Loss}
    \end{subfigure}
    
    \vspace{1em}  

    \begin{subfigure}{0.4\textwidth}
        \centering
        \includegraphics[width=\linewidth]{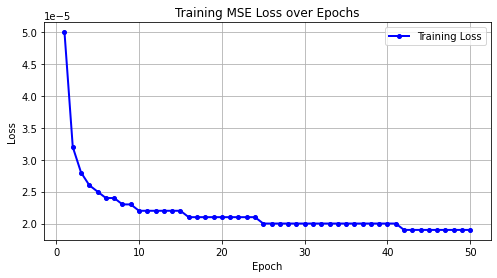}
        \caption{MSE Loss}
    \end{subfigure}
    \hspace{0.03\textwidth}
    \begin{subfigure}{0.4\textwidth}
        \centering
        \includegraphics[width=\linewidth]{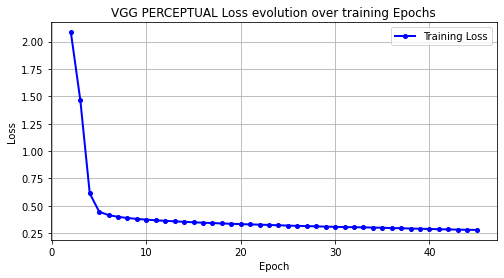}
        \caption{VGG Loss}
    \end{subfigure}
    
    \caption{Evaluation of Loss Functions in DenoMamba: Training Loss Curves for Charbonnier, L1, MSE, and VGG loss}
    \label{fig:loss_evaluation}
\end{figure}

Figure \ref{fig:loss_evaluation} Illustrates the evolution of the LFs considered in this study, including the Charbonnier loss, the L1 loss, the MSE loss and the deep perceptual loss based on VGG during the training epochs. It is worth noticing that Charbonnier loss exhibits a fast convergence over time without significant fluctuations, in contrast to MSE loss, which decreases but eventually plateaus at certain epochs. In comparison, VGG-based deep perceptual loss shows a more gradual and consistent learning process, indicating a smoother convergence.

\begin{figure}[h]
    \centering
    \resizebox{0.8\textwidth}{!}{%
    \begin{minipage}{\textwidth}
        \subfloat[]{%
            \begin{overpic}[width=0.25\textwidth]{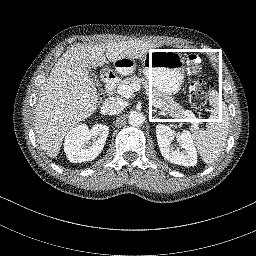}
                \put(55,-10){\adjustbox{frame=1pt, bgcolor=white}{\includegraphics[width=0.15\linewidth]{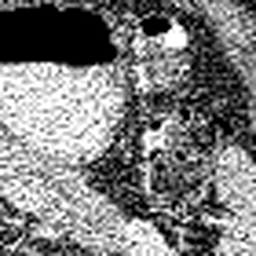}}} 
            \end{overpic}
        }
        \hfill
        \subfloat[]{%
            \begin{overpic}[width=0.25\textwidth]{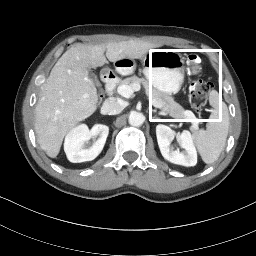}
                \put(55,-10){\adjustbox{frame=1pt, bgcolor=white}{\includegraphics[width=0.15\linewidth]{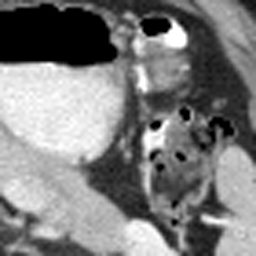}}}
            \end{overpic}
        }
        \hfill
        \subfloat[]{%
            \begin{overpic}[width=0.25\textwidth]{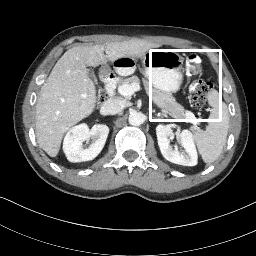}
                \put(55,-10){\adjustbox{frame=1pt, bgcolor=white}{\includegraphics[width=0.15\linewidth]{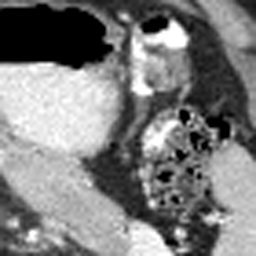}}}
            \end{overpic}
        }

        \subfloat[]{%
            \begin{overpic}[width=0.25\textwidth]{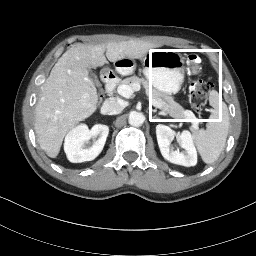}
                \put(55,-10){\adjustbox{frame=1pt, bgcolor=white}{\includegraphics[width=0.15\linewidth]{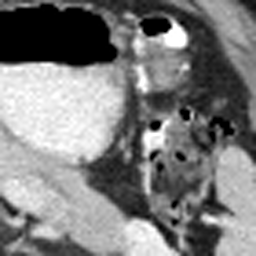}}}
            \end{overpic}
        }
        \hfill
        \subfloat[]{%
            \begin{overpic}[width=0.25\textwidth]{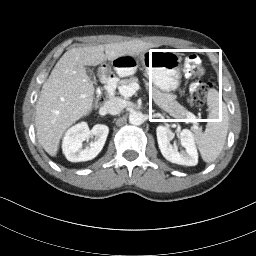}
                \put(55,-10){\adjustbox{frame=1pt, bgcolor=white}{\includegraphics[width=0.15\linewidth]{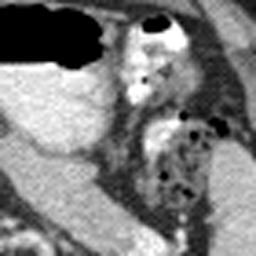}}}
            \end{overpic}
        }
        \hfill
        \subfloat[]{%
            \begin{overpic}[width=0.25\textwidth]{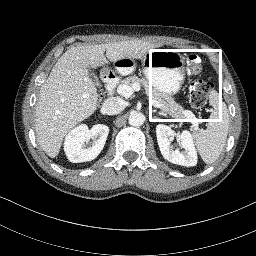}
                \put(55,-10){\adjustbox{frame=1pt, bgcolor=white}{\includegraphics[width=0.15\linewidth]{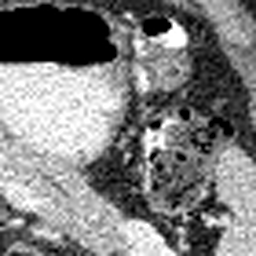}}}
            \end{overpic}
        }    
    \end{minipage}
    }
    \caption{Reconstructed LDCT images using different loss functions. (a) LDCT input, (b-e) reconstructions using Charbonnier, L1, MSE, and VGG-based models, respectively. (f) HDCT reference. The display window is set to [-160, 240] HU.}
    \label{fig:ldct_reconstruction}
\end{figure}

\begin{figure}[H]
    \centering
    \begin{subfigure}{0.45\textwidth} 
        \centering
        \includegraphics[width=\linewidth]{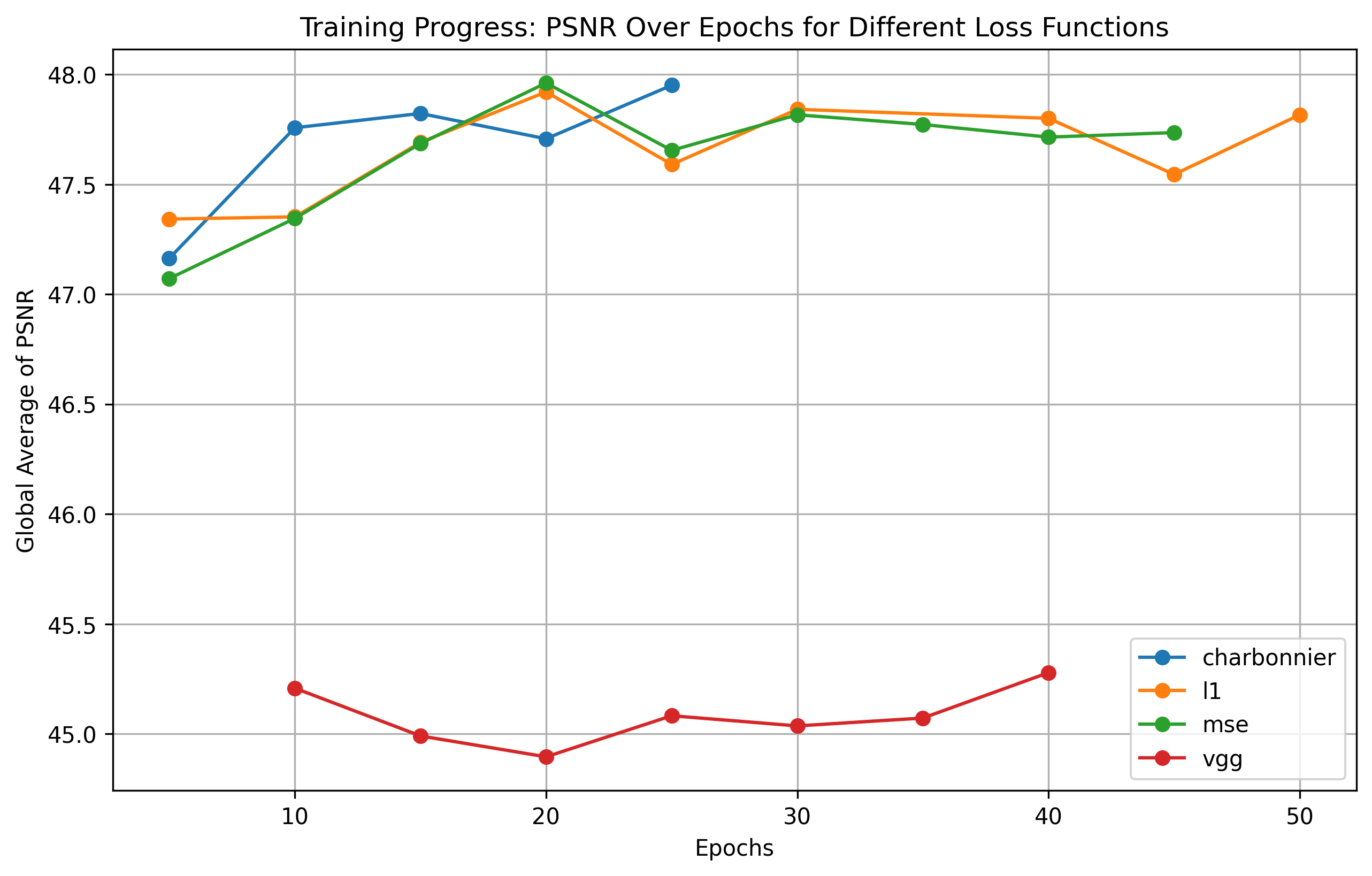}
        \caption{PSNR}
    \end{subfigure}
    \hspace{0.05\textwidth}
    \begin{subfigure}{0.45\textwidth}
        \centering
        \includegraphics[width=\linewidth]{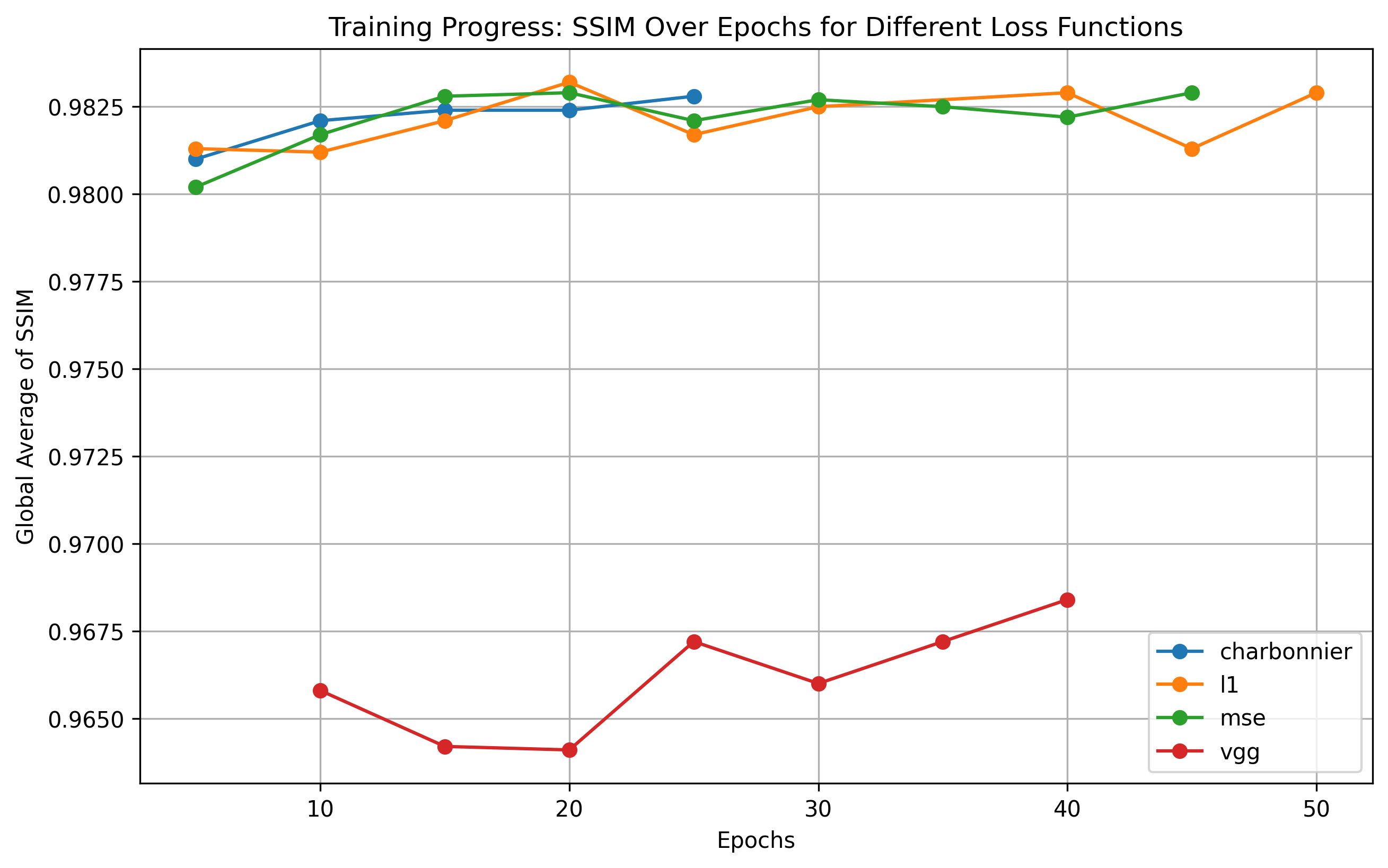}
        \caption{SSIM}
    \end{subfigure}
    \begin{subfigure}{0.45\textwidth}
        \centering
        \includegraphics[width=\linewidth]{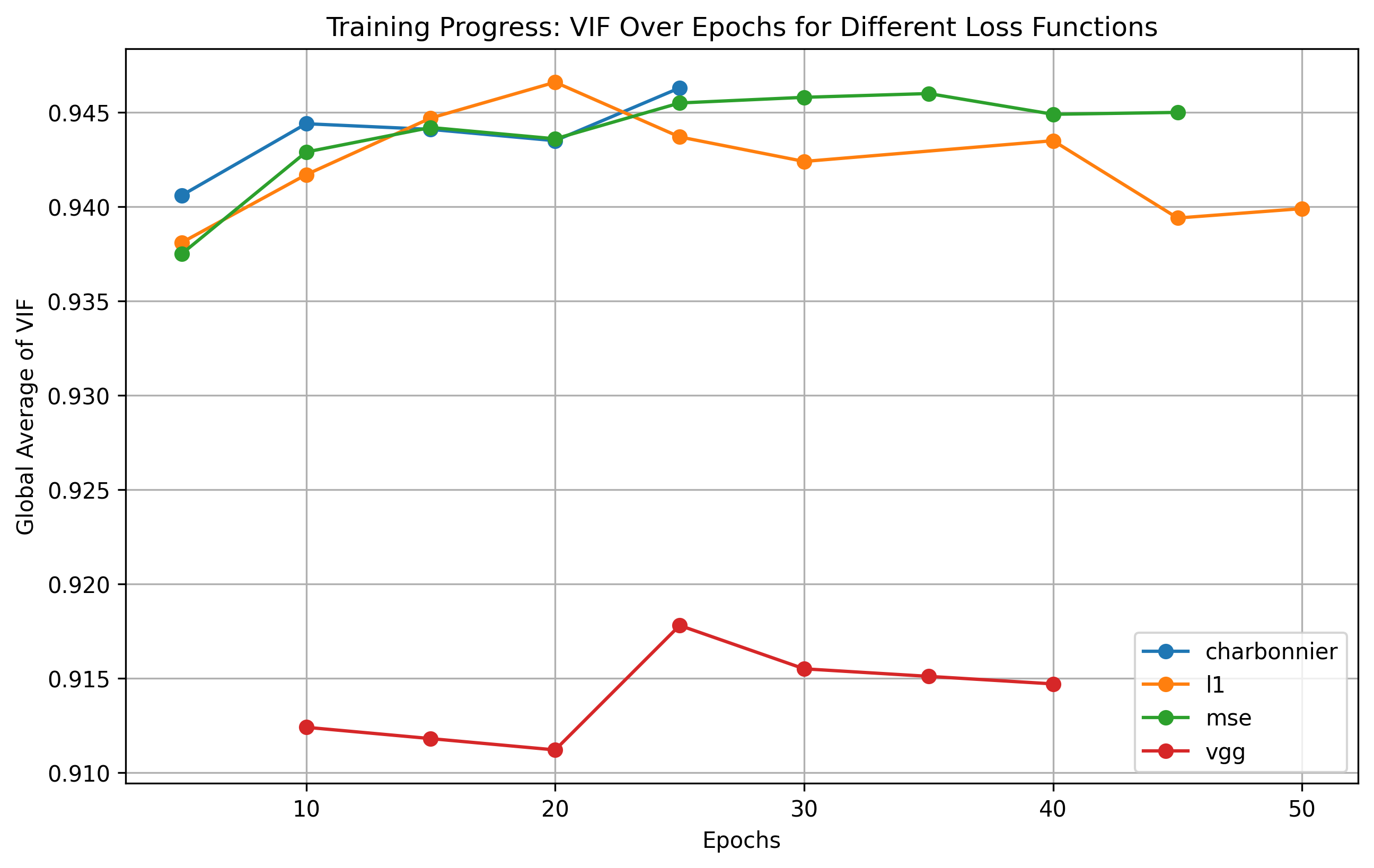}
        \caption{VIF}
    \end{subfigure}
    \hspace{0.05\textwidth}
    \begin{subfigure}{0.45\textwidth}
        \centering
        \includegraphics[width=\linewidth]{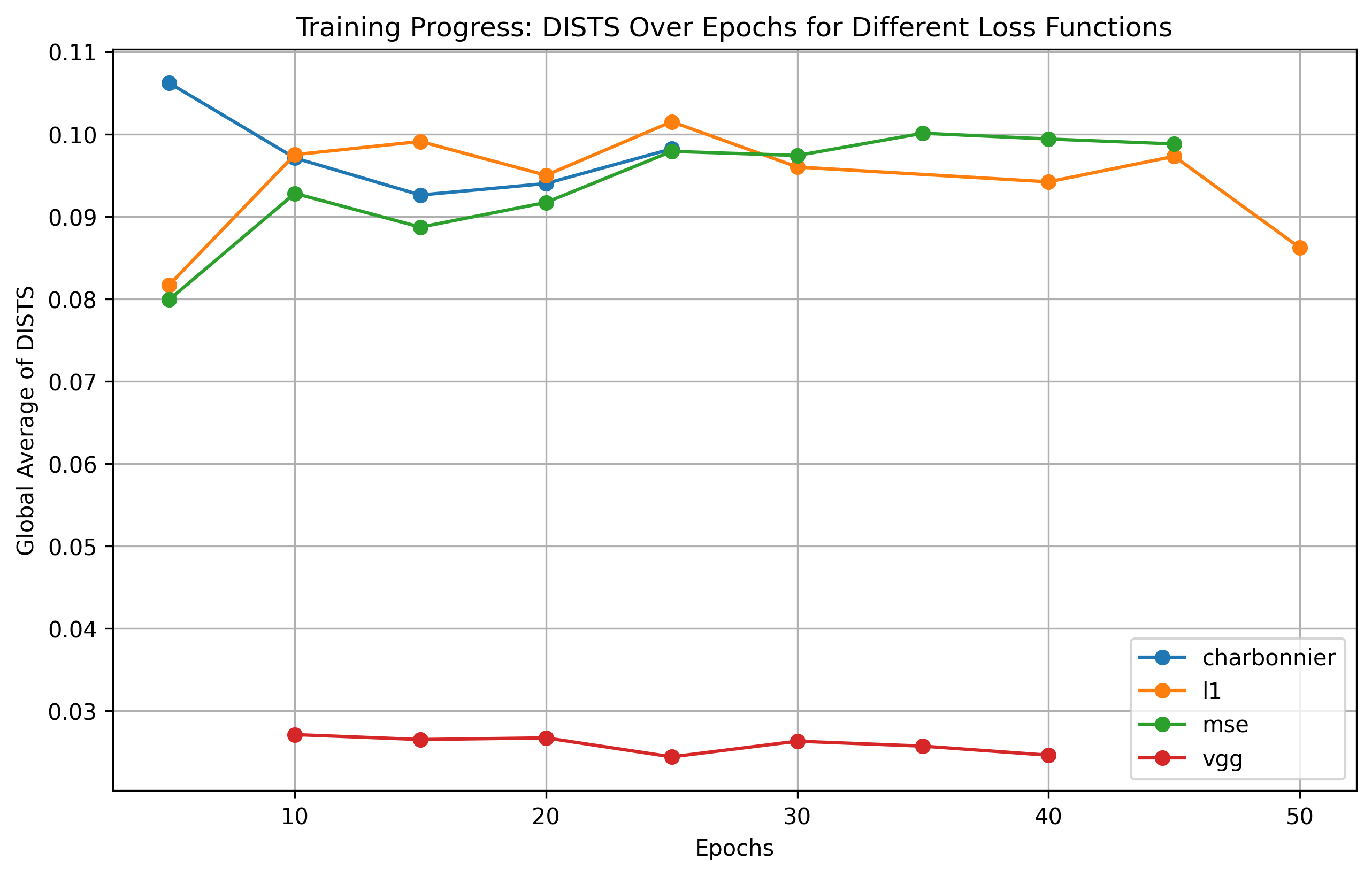}
        \caption{DISTS}
    \end{subfigure}
    \begin{subfigure}{0.45\textwidth}
        \centering
        \includegraphics[width=\linewidth]{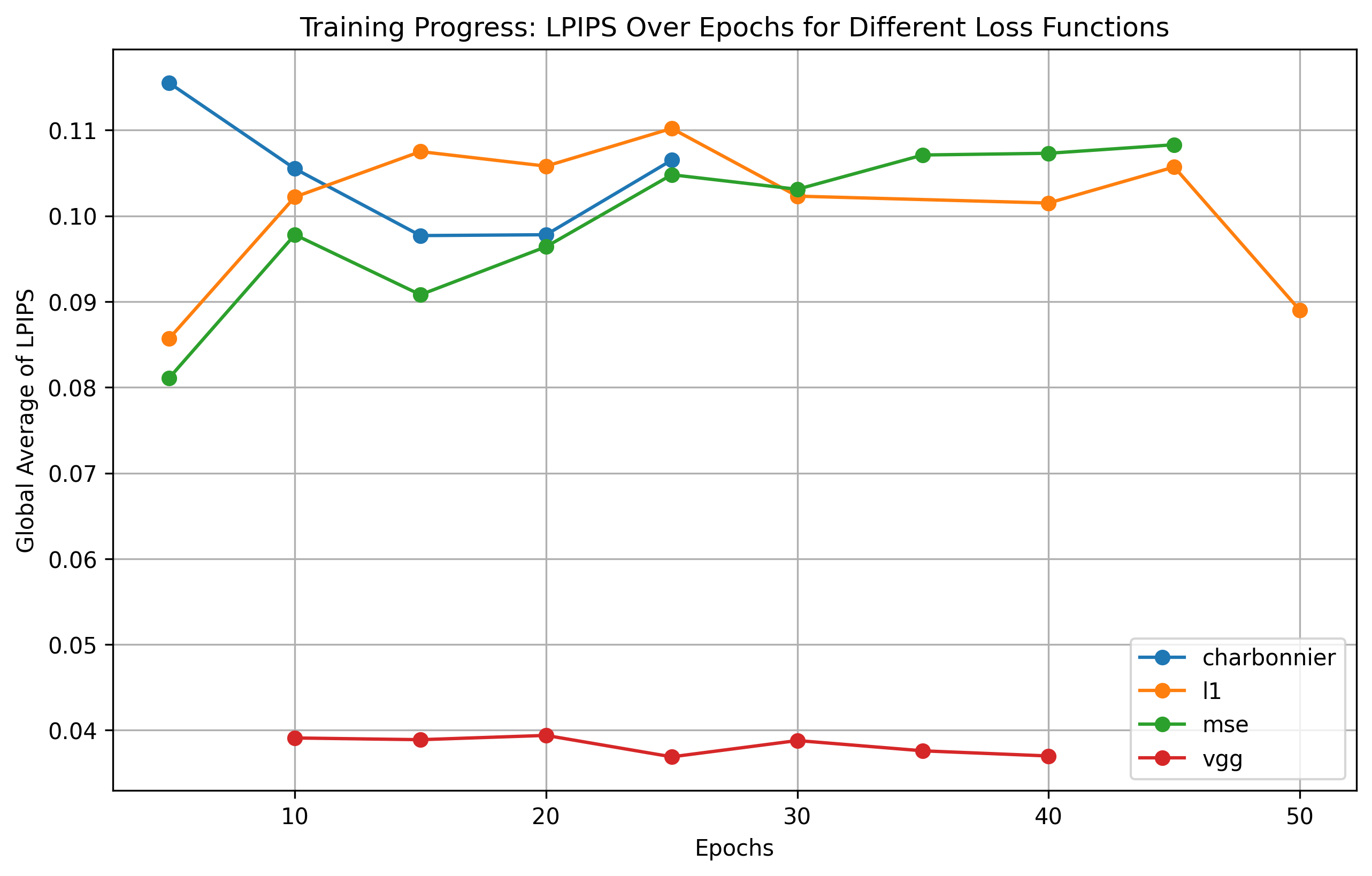}
        \caption{LPIPS}
    \end{subfigure}
    \hspace{0.05\textwidth}
    \begin{subfigure}{0.45\textwidth}
        \centering
        \includegraphics[width=\linewidth]{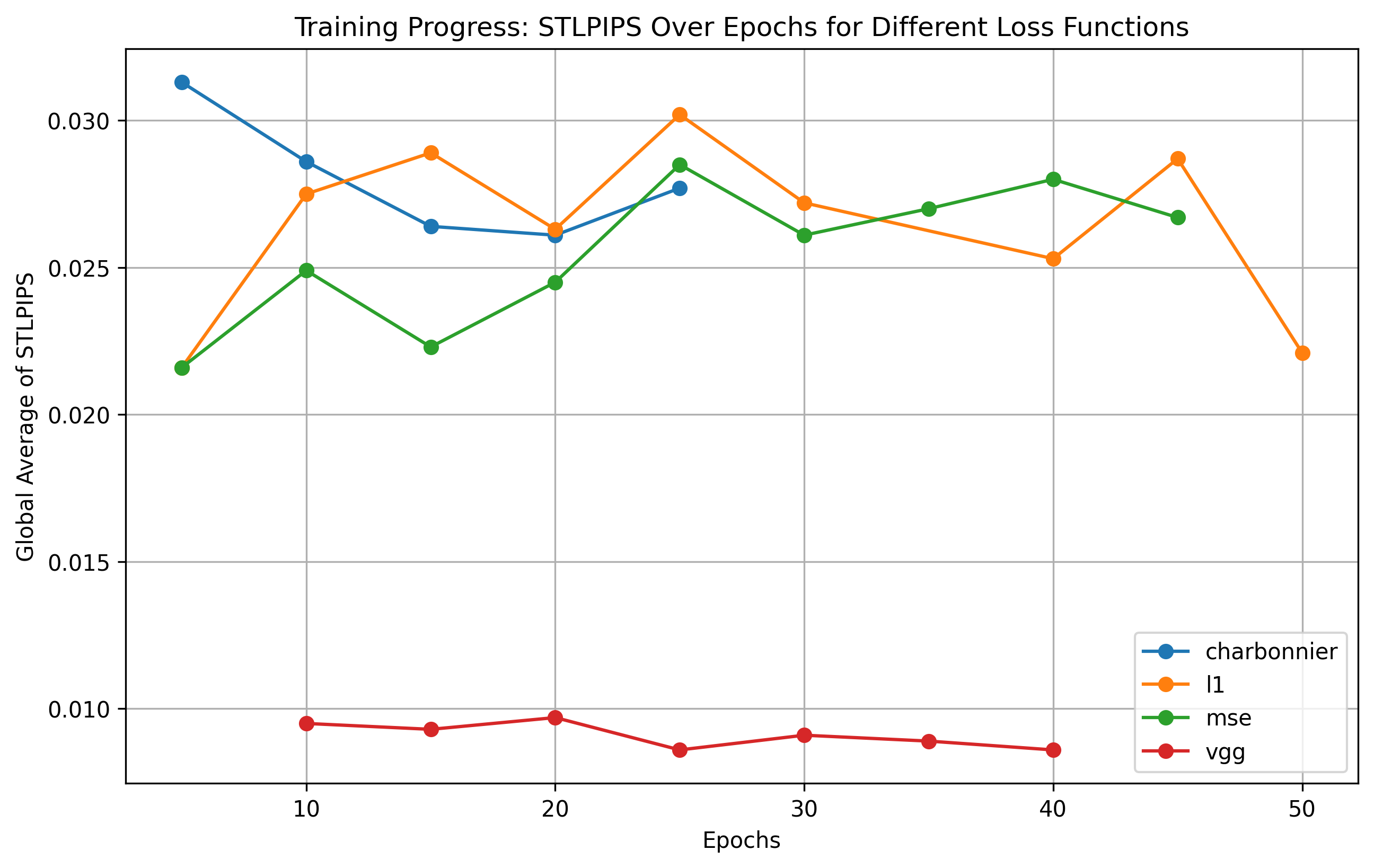}
        \caption{ST-LPIPS}
    \end{subfigure}
    \caption{Training Progress: Full-Reference QA Metrics (PSNR, SSIM, VIF, LPIPS, ST-LPIPS, DISTS)}
    \label{fig:full_ref}
\end{figure}
\begin{figure}[H]
    \centering
    \begin{subfigure}{0.45\textwidth}
        \centering
        \includegraphics[width=\linewidth]{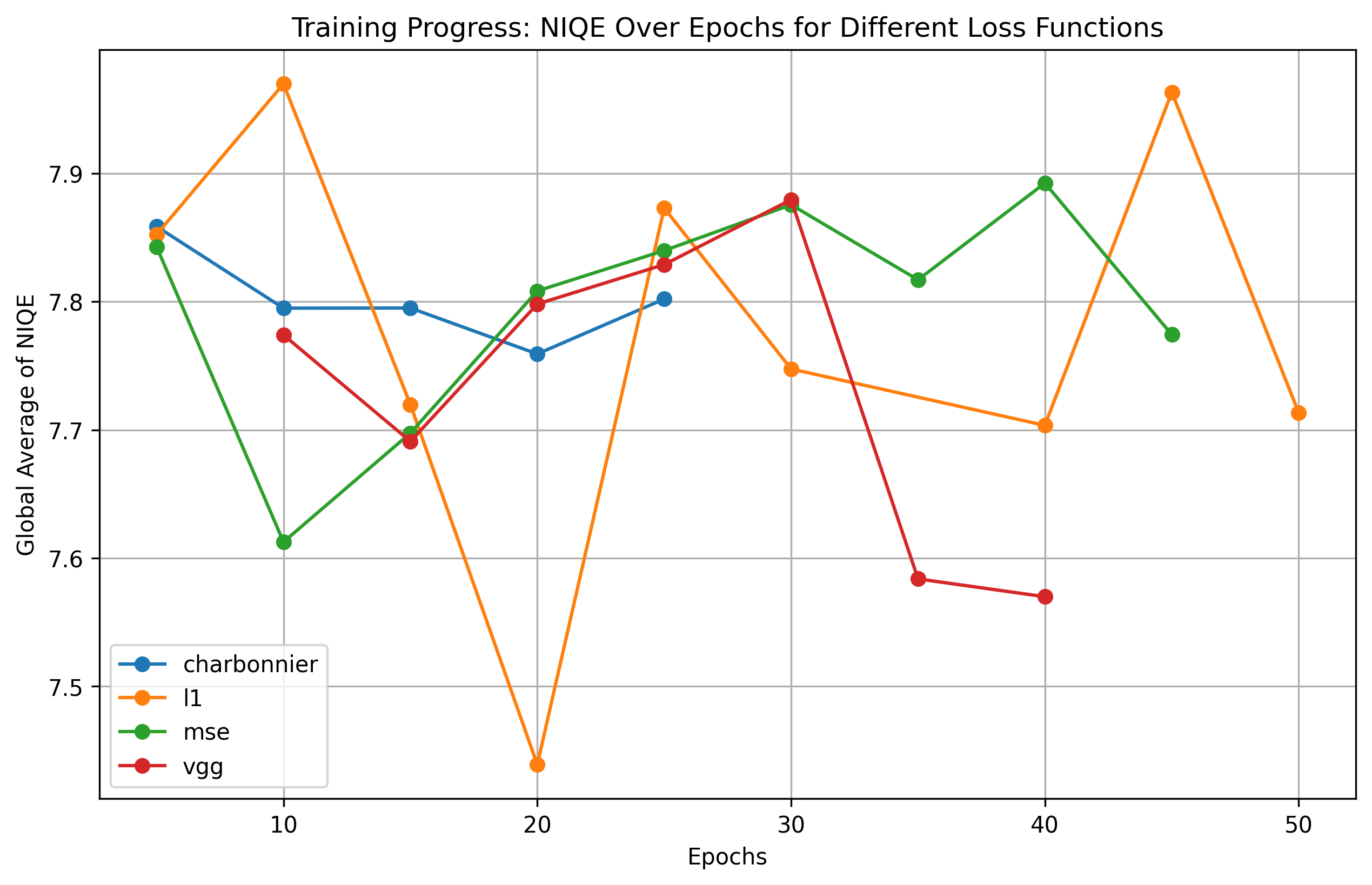}
        \caption{NIQE}
    \end{subfigure}
    \hfill
    \begin{subfigure}{0.45\textwidth}
        \centering
        \includegraphics[width=\linewidth]{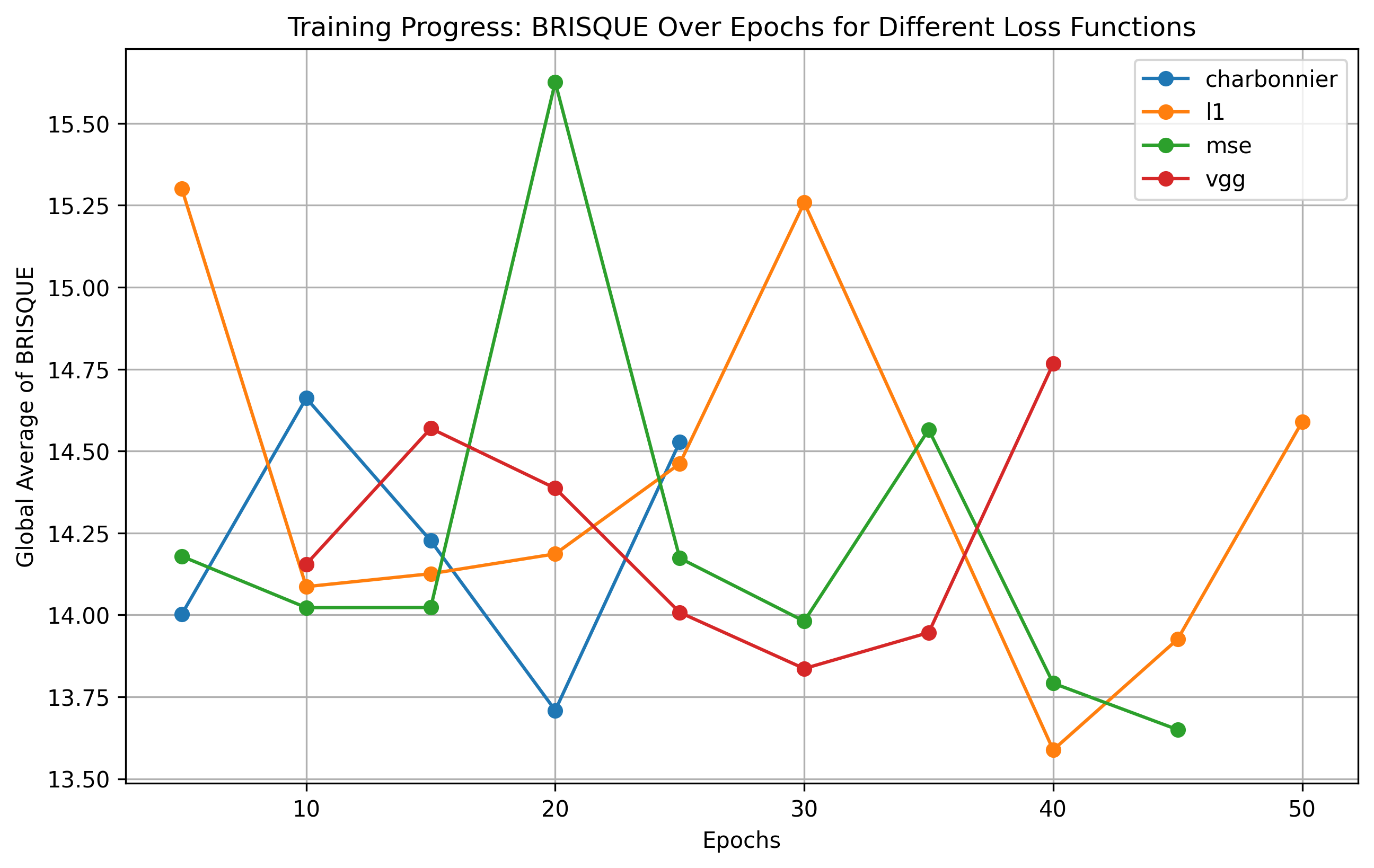}
        \caption{BRISQUE}
    \end{subfigure}
    
    \begin{subfigure}{0.45\textwidth}
        \centering
        \includegraphics[width=\linewidth]{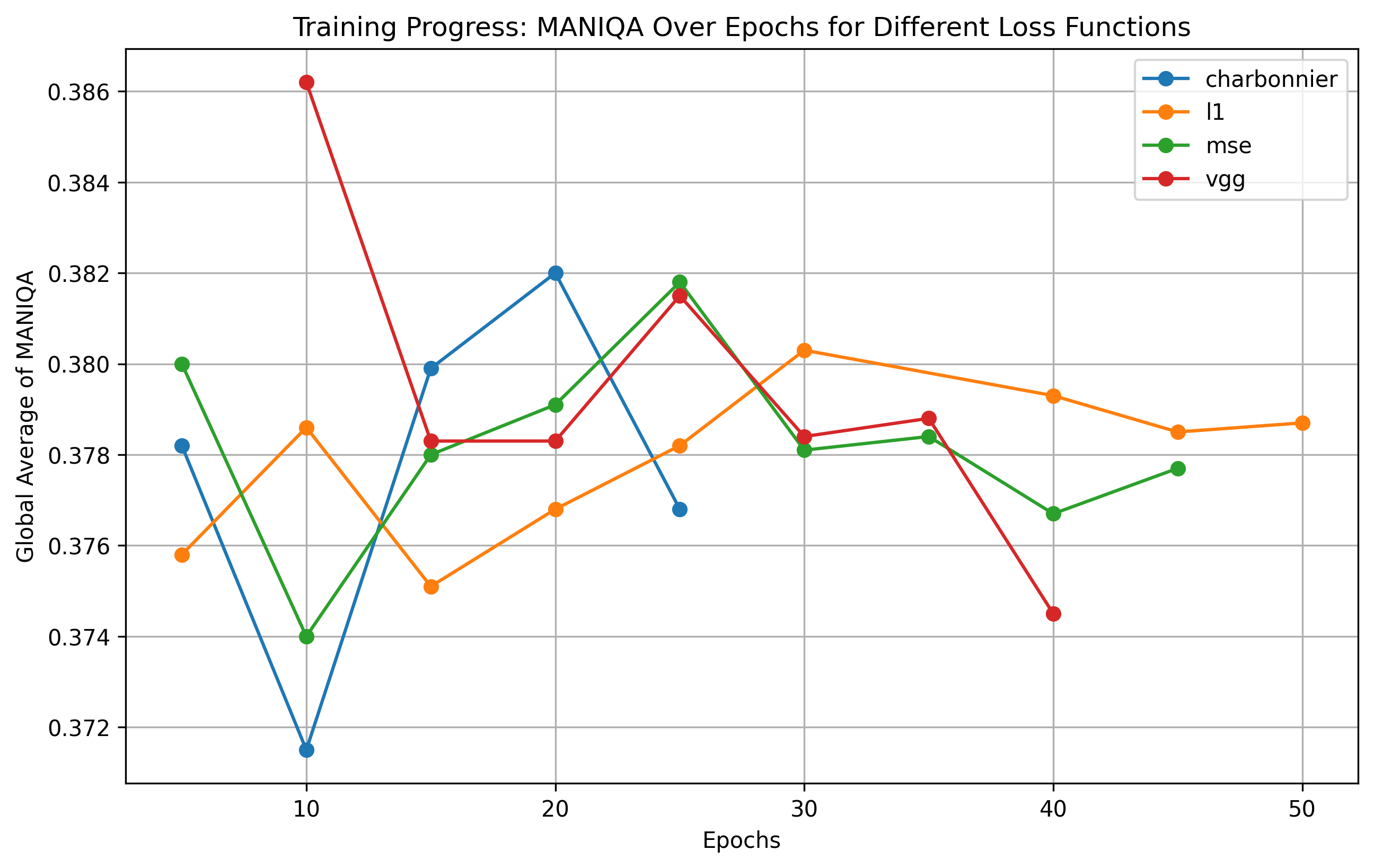}
        \caption{MANIQA}
    \end{subfigure}
    \hfill
    \begin{subfigure}{0.45\textwidth}
        \centering
        \includegraphics[width=\linewidth]{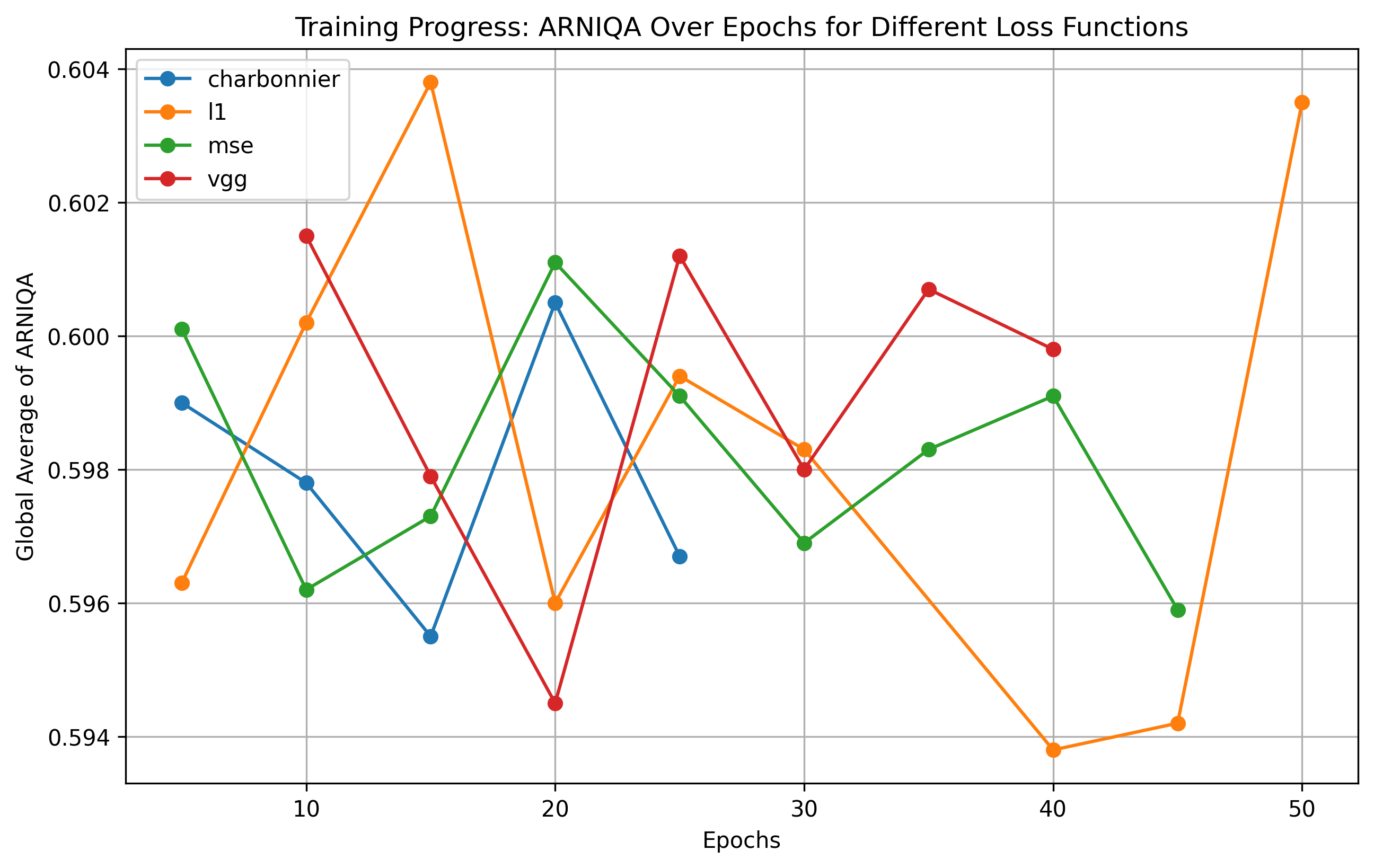}
        \caption{ARNIQA}
    \end{subfigure}
    
    \begin{subfigure}{0.45\textwidth}
        \centering
        \includegraphics[width=\linewidth]{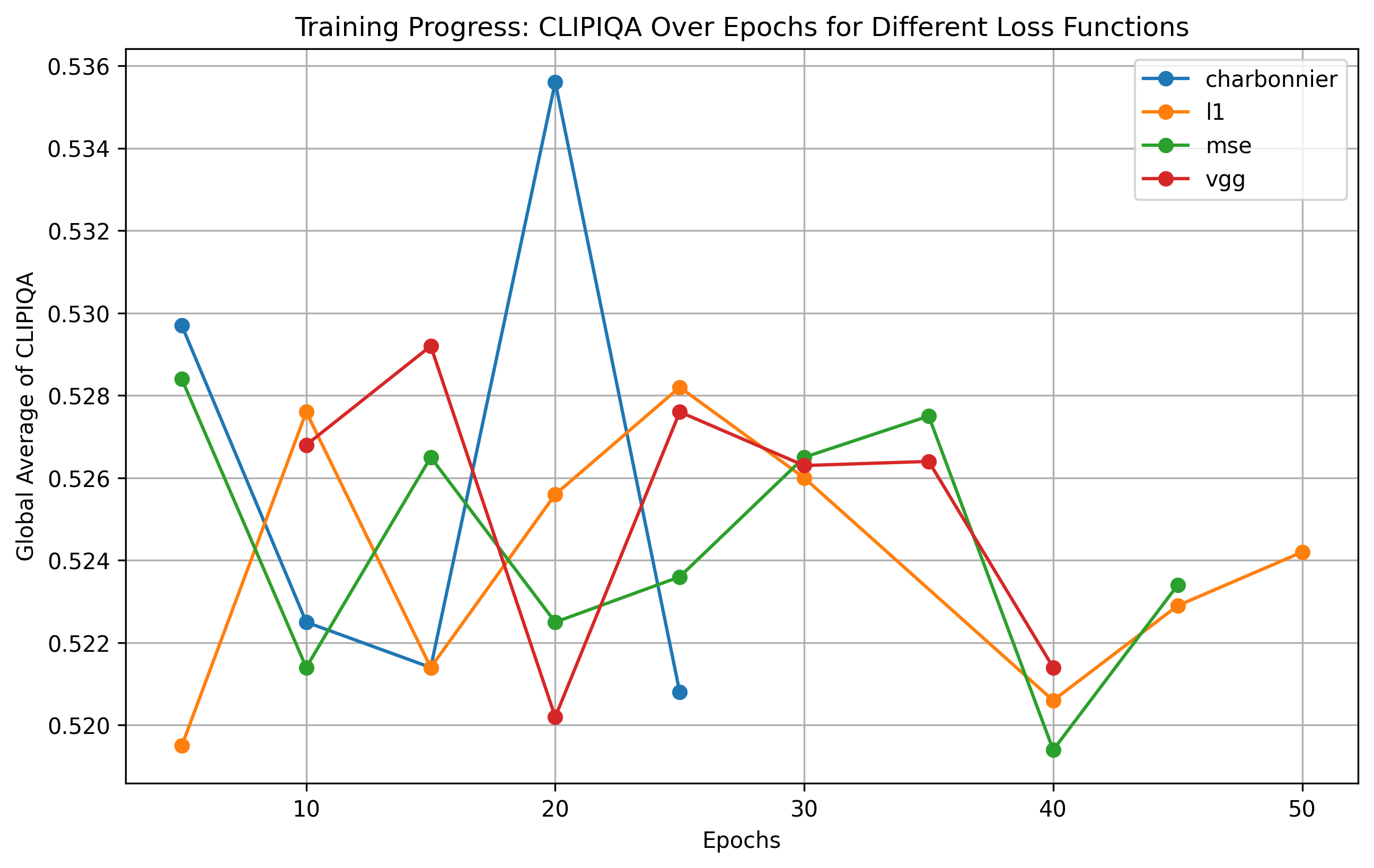}
        \caption{CLIP-IQA}
    \end{subfigure}
    \hfill
    \begin{subfigure}{0.45\textwidth}
        \centering
        \includegraphics[width=\linewidth]{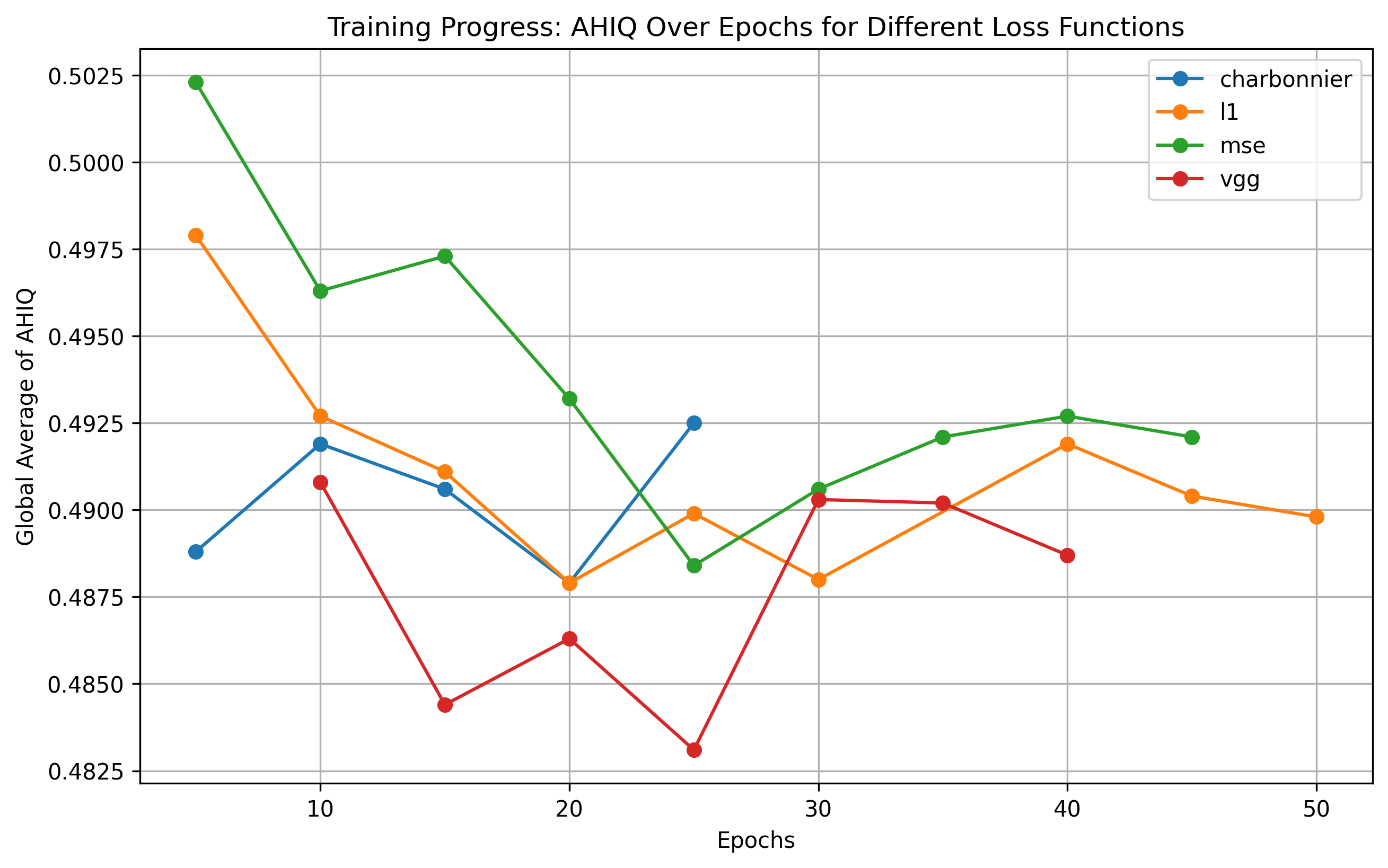}
        \caption{AHIQ}
    \end{subfigure}
    \begin{subfigure}{0.45\textwidth}
        \centering
        \includegraphics[width=\linewidth]{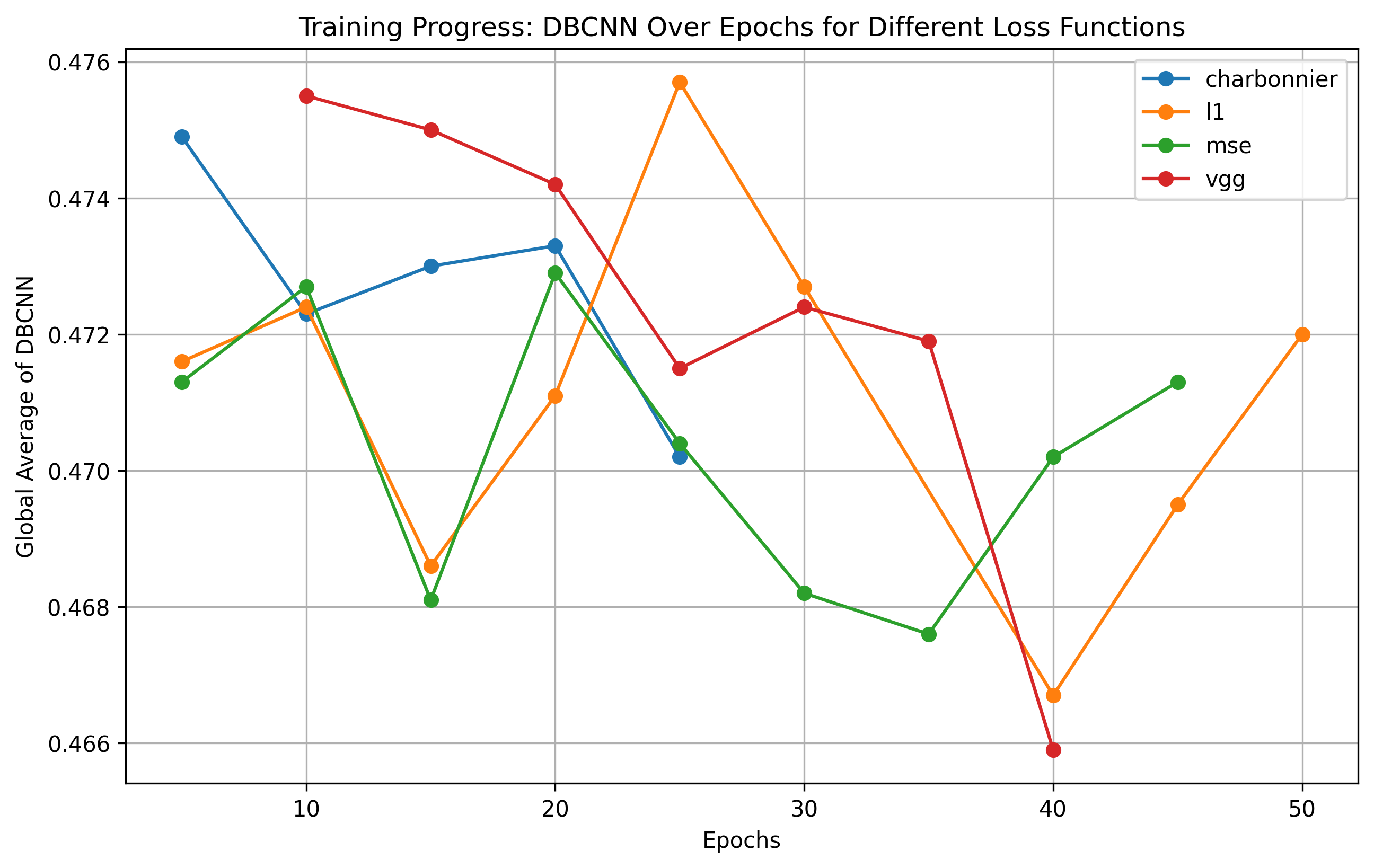}
        \caption{DBCNN}
    \end{subfigure}
    \caption{Training Progress: No-Reference IQA Metrics}
    \label{fig:no_ref}
\end{figure}
Figure \ref{fig:ldct_reconstruction} illustrates the direct impact of LFs on the quality of enhanced LDCT images. Notably, feature-based loss (VGG) preserves fine details and structural textures more effectively than the other pixel-based losses. This observation is further validated by the IQA results in Figure \ref{fig:full_ref}.
Furthermore, our analysis (Figure~\ref{fig:no_ref}) reveals that NR quality metrics, including advanced methods such as CLIP-IQA, exhibit irregular and inconsistent variations that do not align with the behavior of the LFs. This inconsistency suggests that these metrics are unstable across the different LFs considered in the study.

It is important to emphasize that this study does not aim to measure or quantify any form of statistical correlation between IQA metrics and LFs. Instead, we  qualitatively analyze the degree of agreement or disagreement that may exist between the evolution of each IQA metric and the corresponding LF during the evolution of the training process. In other words, the aim is to check whether, during the enhancement process, the evolution of the LF is consistent with that of the image quality metric. A similar study on the stability and coherence of some metrics with respect to the evolution of the level of degradation was carried out in \cite{beghdadi2020critical} in the context of video surveillance \cite{beghdadi2020perceptual}.
As illustrated in Figure \ref{fig:loss_evaluation}, the LFs curves tend to decrease smoothly and monotonically, as expected, while in Figures \ref{fig:full_ref} and \ref{fig:no_ref} most IQA metrics demonstrate chaotic and unstable behavior over training epochs. This highlights a lack of consistency in the evolution of IQA metrics compared to LFs throughout the training process. 
\begin{table}[h]
    \centering
    \begin{tabular}{|l|c|c|c|c|}
        \hline
        IQA Metric & L1 Loss & MSE Loss & Charbonnier Loss & VGG Loss \\
        \hline
        PSNR $\uparrow$    & Moderate & Moderate  & Low & Moderate  \\
        \hline
        SSIM $\uparrow$    & High     & High    & High     & Moderate  \\
        \hline
        LPIPS $\downarrow$ & Moderate & Low     & Moderate  & High \\
        \hline
        ST-LPIPS $\downarrow$ & Low    & Low     & Low  & High  \\
        \hline
        VIF $\uparrow$     & Low      & Moderate & Moderate  & Moderate  \\
        \hline
        DISTS $\downarrow$ & Moderate & Low     & Low  & High \\
        \hline
    \end{tabular}
    \caption{Summary of Consistency Between FR IQA Metrics and Loss Functions}
    \label{tab:consistency}
\end{table}

Table~\ref{tab:consistency} summarizes the observed consistency levels between different LFs and full-reference IQA metrics. The level of consistency (Low, Moderate, High) is based on visual inspection of the curves shown in Figures~\ref{fig:loss_evaluation} and~\ref{fig:full_ref}. For example, when using the VGG perceptual loss, the curves corresponding to perceptual metrics such as LPIPS, ST-LPIPS, and DISTS are smoother and more monotonic, indicating relatively higher consistency. In contrast, the curves of metrics like PSNR, SSIM, and VIF exhibit abrupt changes, particularly between epochs 20 and 25, which suggests moderate or inconsistent alignment. 
Pixel-based LFs such as L1, Charbonnier, and MSE tend to generate more irregular and non-monotonic IQA curves, especially for perceptual metrics like LPIPS and DISTS. This behavior may be attributed to the fact that these LFs operate in the pixel domain, without explicitly modeling perceptual similarity. Conversely, the improved consistency observed with VGG-based LF, for some metrics, likely stems from its optimization in a deep feature space that better aligns with human visual perception.
These findings further support the conclusion that, although VGG loss demonstrates more reliable perceptual consistency with FR perceptual metrics, it does not exhibit overall consistency with the other IQA metrics. In summary, the irregular behavior observed on almost all curves (Fig \ref{fig:full_ref} and \ref{fig:no_ref}) implies
that the considered representative IQA metrics  are globally  insufficient for reliably guiding the training process, as their correlation with LF optimization remains unpredictable.
\section{Conclusion}
Several observations and conclusions can be drawn from this study. The first is that, in view of the results obtained, there is no clear consistency between LFs and image quality measurements, which are fairly representative of the state-of-the-art considered in this study. We can also notice the absence of an indicator for choosing a quality metric to control the iterative enhancement process. In fact, according to our analysis, there is no coherent link between the quality level and the evolution of LFs. For example, a low LF value does not necessarily correspond to a high image quality level. We can also conclude that there is still room for the development of more plausible quality metrics and LFs, particularly in the field of image quality enhancement. One way to achieve this objective is to design LFs that are better aligned with human perception by integrating recent advances in IQA metrics considering, for example, visual attention mechanisms and patch-based approaches. This study opens new perspectives for the development of more consistent and quality-guided LFs in order to optimize deep learning-based models for image quality enhancement.

%
%
%
\bibliographystyle{splncs04}
\bibliography{mybibliography}

\begin{thebibliography}{10}
\providecommand{\url}[1]{\texttt{#1}}
\providecommand{\urlprefix}{URL }
\providecommand{\doi}[1]{https://doi.org/#1}

\bibitem{agnolucci2024arniqa}
Agnolucci, L., Galteri, L., Bertini, M., Del~Bimbo, A.: Arniqa: Learning distortion manifold for image quality assessment. In: Proceedings of the IEEE/CVF Winter Conference on Applications of Computer Vision. pp. 189--198 (2024)

\bibitem{beghdadi2020perceptual}
Beghdadi, A., Bezzine, I., Qureshi, M.A.: A perceptual quality-driven video surveillance system. In: 2020 IEEE 23rd International Multitopic Conference (INMIC). pp.~1--6. IEEE (2020)

\bibitem{beghdadi2020critical}
Beghdadi, A., Qureshi, M.A., Amirshahi, S.A., Chetouani, A., Pedersen, M.: A critical analysis on perceptual contrast and its use in visual information analysis and processing. IEEE Access  \textbf{8},  156929--156953 (2020)

\bibitem{chen2017low}
Chen, H., Zhang, Y., Kalra, M.K., Lin, F., Chen, Y., Liao, P., Zhou, J., Wang, G.: Low-dose ct with a residual encoder-decoder convolutional neural network. IEEE transactions on medical imaging  \textbf{36}(12),  2524--2535 (2017)

\bibitem{chen2024low}
Chen, Z., Chen, T., Wang, C., Gao, Q., Niu, C., Wang, G., Shan, H.: Low-dose ct denoising with language-engaged dual-space alignment. In: 2024 IEEE International Conference on Bioinformatics and Biomedicine (BIBM). pp. 3088--3091. IEEE (2024)

\bibitem{chen2023ascon}
Chen, Z., Gao, Q., Zhang, Y., Shan, H.: Ascon: Anatomy-aware supervised contrastive learning framework for low-dose ct denoising. In: International Conference on Medical Image Computing and Computer-Assisted Intervention. pp. 355--365. Springer (2023)

\bibitem{chen2024lit}
Chen, Z., Niu, C., Gao, Q., Wang, G., Shan, H.: Lit-former: Linking in-plane and through-plane transformers for simultaneous ct image denoising and deblurring. IEEE Transactions on Medical Imaging  \textbf{43}(5),  1880--1894 (2024)

\bibitem{9298952}
Ding, K., Ma, K., Wang, S., Simoncelli, E.P.: Image quality assessment: Unifying structure and texture similarity. IEEE Transactions on Pattern Analysis and Machine Intelligence  \textbf{44}(5),  2567--2581 (2022). \doi{10.1109/TPAMI.2020.3045810}

\bibitem{gao2022cocodiff}
Gao, Q., Shan, H.: Cocodiff: a contextual conditional diffusion model for low-dose ct image denoising. In: Developments in X-Ray Tomography XIV. vol. 12242, pp. 92--98. SPIE (2022)

\bibitem{ge2020adaptive}
Ge, Y., Su, T., Zhu, J., Deng, X., Zhang, Q., Chen, J., Hu, Z., Zheng, H., Liang, D.: Adaptive-net: deep computed tomography reconstruction network with analytical domain transformation knowledge. Quantitative imaging in medicine and surgery  \textbf{10}(2), ~415 (2020)

\bibitem{ghildyal2022stlpips}
Ghildyal, A., Liu, F.: Shift-tolerant perceptual similarity metric. In: European Conference on Computer Vision (2022)

\bibitem{gholizadeh2020deep}
Gholizadeh-Ansari, M., Alirezaie, J., Babyn, P.: Deep learning for low-dose ct denoising using perceptual loss and edge detection layer. Journal of digital imaging  \textbf{33},  504--515 (2020)

\bibitem{huang2021gan}
Huang, Z., Zhang, J., Zhang, Y., Shan, H.: Du-gan: Generative adversarial networks with dual-domain u-net-based discriminators for low-dose ct denoising. IEEE Transactions on Instrumentation and Measurement  \textbf{71},  1--12 (2021)

\bibitem{kulathilake2023review}
Kulathilake, K.S.H., Abdullah, N.A., Sabri, A.Q.M., Lai, K.W.: A review on deep learning approaches for low-dose computed tomography restoration. Complex \& Intelligent Systems  \textbf{9}(3),  2713--2745 (2023)

\bibitem{9857006}
Lao, S., Gong, Y., Shi, S., Yang, S., Wu, T., Wang, J., Xia, W., Yang, Y.: Attentions help cnns see better: Attention-based hybrid image quality assessment network. In: 2022 IEEE/CVF Conference on Computer Vision and Pattern Recognition Workshops (CVPRW). pp. 1139--1148 (2022). \doi{10.1109/CVPRW56347.2022.00123}

\bibitem{le1992image}
Le~N{\'e}grate, A., Beghdadi, A., Dupoisot, H.: An image enhancement technique and its evaluation through bimodality analysis. CVGIP: Graphical Models and Image Processing  \textbf{54}(1),  13--22 (1992)

\bibitem{li2024ct}
Li, L., Wei, W., Yang, L., Zhang, W., Dong, J., Zhao, W.: Ct-mamba: A hybrid convolutional state space model for low-dose ct denoising. arXiv preprint arXiv:2411.07930  (2024)

\bibitem{li2021low}
Li, Z., Shi, W., Xing, Q., Miao, Y., He, W., Yang, H., Jiang, Z.: Low-dose ct image denoising with improving wgan and hybrid loss function. Computational and Mathematical Methods in Medicine  \textbf{2021}(1),  2973108 (2021)

\bibitem{liang2020edcnn}
Liang, T., Jin, Y., Li, Y., Wang, T.: Edcnn: Edge enhancement-based densely connected network with compound loss for low-dose ct denoising. In: 2020 15th IEEE International conference on signal processing (ICSP). vol.~1, pp. 193--198. IEEE (2020)

\bibitem{liu2023solving}
Liu, W., Ding, H.: Solving low-dose ct reconstruction via gan with local coherence. In: International Conference on Medical Image Computing and Computer-Assisted Intervention. pp. 524--534. Springer (2023)

\bibitem{mazandarani2023unext}
Mazandarani, F.N., Babyn, P., Alirezaie, J.: Unext: a low-dose ct denoising unet model with the modified convnext block. In: ICASSP 2023-2023 IEEE International Conference on Acoustics, Speech and Signal Processing (ICASSP). pp.~1--5. IEEE (2023)

\bibitem{mccollough2016tu}
McCollough, C.: Tu-fg-207a-04: overview of the low dose ct grand challenge. Medical physics  \textbf{43}(6Part35),  3759--3760 (2016)

\bibitem{mittal2012making}
Mittal, A., Soundararajan, R., Bovik, A.C.: Making a “completely blind” image quality analyzer. IEEE Signal processing letters  \textbf{20}(3),  209--212 (2012)

\bibitem{moorthy2011blind}
Moorthy, A.K., Bovik, A.C.: Blind image quality assessment: From natural scene statistics to perceptual quality. IEEE transactions on Image Processing  \textbf{20}(12),  3350--3364 (2011)

\bibitem{ozturk2024denomamba}
{\"O}zt{\"u}rk, {\c{S}}., Duran, O.C., {\c{C}}ukur, T.: Denomamba: A fused state-space model for low-dose ct denoising. arXiv preprint arXiv:2409.13094  (2024)

\bibitem{sheikh2005visual}
Sheikh, H.R., Bovik, A.C.: A visual information fidelity approach to video quality assessment. In: The first international workshop on video processing and quality metrics for consumer electronics. vol.~7, pp. 2117--2128. sn (2005)

\bibitem{wang2023ctformer}
Wang, D., Fan, F., Wu, Z., Liu, R., Wang, F., Yu, H.: Ctformer: convolution-free token2token dilated vision transformer for low-dose ct denoising. Physics in Medicine \& Biology  \textbf{68}(6),  065012 (2023)

\bibitem{wang2023exploring}
Wang, J., Chan, K.C., Loy, C.C.: Exploring clip for assessing the look and feel of images. In: Proceedings of the AAAI conference on artificial intelligence. vol.~37, pp. 2555--2563 (2023)

\bibitem{wang2022exploring}
Wang, J., Chan, K.C., Loy, C.C.: Exploring clip for assessing the look and feel of images. In: AAAI (2023)

\bibitem{wang2023self}
Wang, Z., Liu, M., Cheng, X., Zhu, J., Wang, X., Gong, H., Liu, M., Xu, L.: Self-adaption and texture generation: A hybrid loss function for low-dose ct denoising. Journal of Applied Clinical Medical Physics  \textbf{24}(9),  e14113 (2023)

\bibitem{wang2004image}
Wang, Z., Bovik, A.C., Sheikh, H.R., Simoncelli, E.P.: Image quality assessment: from error visibility to structural similarity. IEEE transactions on image processing  \textbf{13}(4),  600--612 (2004)

\bibitem{xia2022low}
Xia, W., Lyu, Q., Wang, G.: Low-dose ct using denoising diffusion probabilistic model for 20 $\times$ speedup. arXiv preprint  \textbf{arXiv:2209.15136} (2022)

\bibitem{xiong2024re}
Xiong, L., Li, N., Qiu, W., Luo, Y., Li, Y., Zhang, Y.: Re-unet: a novel multi-scale reverse u-shape network architecture for low-dose ct image reconstruction. Medical \& Biological Engineering \& Computing  \textbf{62}(3),  701--712 (2024)

\bibitem{yang2018low}
Yang, Q., Yan, P., Zhang, Y., Yu, H., Shi, Y., Mou, X., Kalra, M.K., Zhang, Y., Sun, L., Wang, G.: Low-dose ct image denoising using a generative adversarial network with wasserstein distance and perceptual loss. IEEE transactions on medical imaging  \textbf{37}(6),  1348--1357 (2018)

\bibitem{yang2022maniqa}
Yang, S., Wu, T., Shi, S., Lao, S., Gong, Y., Cao, M., Wang, J., Yang, Y.: Maniqa: Multi-dimension attention network for no-reference image quality assessment. In: Proceedings of the IEEE/CVF Conference on Computer Vision and Pattern Recognition. pp. 1191--1200 (2022)

\bibitem{zhang2023novel}
Zhang, J., Niu, Y., Shangguan, Z., Gong, W., Cheng, Y.: A novel denoising method for ct images based on u-net and multi-attention. Computers in Biology and Medicine  \textbf{152},  106387 (2023)

\bibitem{zhang2018unreasonable}
Zhang, R., Isola, P., Efros, A.A., Shechtman, E., Wang, O.: The unreasonable effectiveness of deep features as a perceptual metric. In: Proceedings of the IEEE conference on computer vision and pattern recognition. pp. 586--595 (2018)

\bibitem{zhang2020blind}
Zhang, W., Ma, K., Yan, J., Deng, D., Wang, Z.: Blind image quality assessment using a deep bilinear convolutional neural network. IEEE Transactions on Circuits and Systems for Video Technology  \textbf{30}(1),  36--47 (2020)

\end{thebibliography}
%




\end{document}